\documentclass[11pt,a4paper]{article}
\usepackage[]{acl}

\usepackage{times}
\usepackage{latexsym}

\newcommand{\maxs}{$max_s$ }
\usepackage{subcaption}
\usepackage{float}

\usepackage{verbatimbox}

\usepackage{microtype}

\usepackage{amssymb}
\usepackage{amsmath}
\usepackage{tikz, tikz-qtree}
\usepackage{todonotes}
\usepackage{booktabs}
\usepackage{multirow}
\usepackage{pdfpages}

\usepackage{ulem}

\title{Probing for Constituency Structure in Neural Language Models}

\author{David Arps$^\dagger$ \qquad 
	Younes Samih$^\dagger$ \qquad 
	Laura Kallmeyer$^\dagger$ \qquad
	Hassan Sajjad$^\ddagger$\\
$^\dagger$Heinrich Heine 
	University D\"usseldorf, 
	D\"usseldorf, Germany \\
	 $^\ddagger$Qatar Computing Research Institute,  
  Hamad Bin Khalifa University, Qatar\\
	\texttt{$\{$david.arps,younes.samih,laura.kallmeyer$\}$@hhu.de} \\\quad {\tt hsajjad@hbku.edu.qa}\\~\\
 }

\begin{document}

\maketitle
\begin{abstract}
In this paper, we investigate to which extent contextual neural language models (LMs) implicitly learn syntactic structure. More concretely, we focus on constituent structure as represented in the Penn Treebank (PTB). 
Using standard probing techniques based on diagnostic classifiers, we assess the accuracy of representing constituents of different categories within the neuron activations of a LM such as RoBERTa. In order to make sure that our probe focuses on syntactic knowledge and not on implicit semantic generalizations, we also experiment on a PTB version that is obtained by randomly replacing constituents with each other while keeping syntactic structure, i.e., a semantically ill-formed but syntactically well-formed version of the PTB. We find that 4 pretrained transfomer LMs obtain high performance on our probing tasks even on manipulated data, suggesting that semantic and syntactic knowledge in their representations can be separated and that constituency information is in fact learned by the LM. Moreover, we show that a complete constituency tree can be linearly separated from LM representations.\footnote{Code for our experiments is available at \url{https://github.com/davidarps/constptbprobing}
     }
     
\end{abstract}

%

\section{Introduction}

Over the last years, neural language models (LMs), such as BERT \cite{devlin-etal-2019-bert}, XLNet \cite{xlnet-NEURIPS2019_dc6a7e65}, RoBERTa \citep{liu2019roberta}, and DistilBERT \citep{sanh2020distilbert}, have delivered unmatched results in multiple key Natural Language Processing (NLP) benchmarks \cite{glue:2018, NEURIPS2019-4496bf24}.
Despite the impressive performance, the black-box nature of these models makes it difficult to ascertain whether they implicitly learn to encode linguistic structures, such as constituency or dependency trees. 

There has been a considerable amount of research conducted on questioning which types of linguistic structure are learned by LMs~\citep{tenney:2019:what,conneau:2018:probing,liu-etal-2019-linguistic}. 
The motivation behind asking this question is two-fold. On the one hand, we want to better understand how pre-trained LMs 
solve certain NLP tasks, i.e., how their input features and neuron activations contribute to a specific classification success. 
A second motivation is an interest in distributional evidence for linguistic theory. That is, we are interested in assessing which types of linguistic categories emerge when training a contextual language model, i.e., when training a model only on unlabeled text. The research in this paper is primarily motivated by this second aspect, focusing on syntactic structure, more concretely on constituency structure. We investigate, for instance, for pairs of tokens in a sentence whether a LM implicitly learns which constituent (\texttt{NP}, \texttt{VP}, \dots) the two tokens belong to as their lowest common ancestor (\texttt{LCA}). We use English Penn Treebank data~\citep[PTB, ][]{marcus-etal-1993-building} to conduct experiments.

A number of studies have probed LMs for dependency structure~\citep{hewitt-manning-2019-structural,chen2021probing} and constituency structure~\citep{tenney:2019:what}. We probe constituency structure for the following reasons. 
In contrast to dependency structure, it can be richer concerning the represented abstract syntactic information, since it directly assigns categories to groups of tokens. On the other hand, not all dependency labels are represented in a standard constituency structure; but they can be incorporated as extensions of the corresponding non-terminal nodes (see, e.g., the PTB label \texttt{NP-SBJ} in App.~\ref{app:const:vs:dep}).
To quantify the gain that we get from probing constituency rather than dependency trees, we compare the unlabeled bracketings in the syntactic trees in both formalisms on the PTB \cite{marcus-etal-1993-building, de-marneffe-etal-2006-generating}, where an unlabeled bracketing is the yield of a subtree. 
We find that while $97\%$ of the bracketings in a dependency tree are also present in a constituency tree, only $54\%$ of the bracketings in the constituency tree are present in the dependency tree. 
This shows that constituent trees contain much more fine-grained hierarchical information than dependency trees. 
A further reason for focusing on constituency structure is that this is the type of structure most linguistic theories use.

We use diagnostic classifiers~\cite{hupkes:2018:vis} and perform model-level, layer-level and neuron-level analyses. Most work on diagnostic classifiers performs mean pool over representations when probing for a relationship between two words~\cite{durrani-individual:emnlp20}. We empirically show that mean pool results in lossy representation, and we recommend concatenation of representations as a better way to probe for relations between words. 

A difficulty when probing a LM for whether certain categories are learned is that we cannot be sure that the LM does not learn a different category instead that is also predictive for the category we are interested in. More concretely, when probing for syntax, one should make sure that it is not semantics that one finds and considers to be syntax (since semantic relations influence syntactic structure). This point was also observed by \citet{gulordava-etal-2018-colorless} and \citet{hall-maudslay-cotterell-2021-syntactic}. Therefore, 
we manipulate our data by replacing a subset of tokens with other tokens that appear in similar syntactic contexts, thereby obtaining nonsensical text that still has a reasonable syntactic structure. We then conduct a series of experiments that show that even for these nonsensical sentences, contextual LMs implicitly represent constituency structure. 
Lastly, we tested whether a full syntactic tree can be reconstructed using the linear probe. We achieve a labeled F1 score of 82.6\% for RoBERTa when probing on the non-manipulated dataset in comparison to 51.4\% with a random representation baseline.
The contributions of our work are as follows:
\begin{itemize}
    \item We find that constituency structure is linearly separable at various granularity levels: At the model level, we find that four different LMs achieve similar overall performance on our syntactic probing tasks, but make slightly different predictions. At the layer level, the middle layers are most important. At the neuron level, syntax is heavily distributed across neurons.
    \item We use perturbed data to separate the effect of semantics when probing for syntax, and we find that different sets of neurons capture syntactic and semantic information
    \item We show that a simple linear probe is effective in analyzing representations for syntactic properties and we show that a full constituency tree can be linearly separated from LM representations
\end{itemize}

The rest of the paper is structured as follows.
The next section introduces related work. 
We define our linguistic probing tasks in Sec.~\ref{sec:diagnoseconst}.
Sec.~\ref{sec:methods} introduces our experimental methodology.
Sec.~\ref{sec:expsetup},  \ref{sec:results}, and \ref{sec:tree-reconstruction} discuss our experiments and their results, and Sec.~\ref{sec:conclusion}
 concludes. 

\section{Related work}\label{sec:relatedwork}

\paragraph{Syntactic information in neural LMs}
An important recent line of research 
\cite{adi:2016:fine,hupkes:2018:vis,zhang:bowman:2018:language,blevins:etal:2018:deep,hewitt-manning-2019-structural,hewitt-liang-2019-designing,coenen2019visualizing,tenney:etal:2019:bert,manning2020emergent,hall:maudslay:etal:2020:tale,li:etal:2020:branching,newman:etal:2021:refining,hewitt2021conditional}
has focused on finding latent hierarchical structures encoded in neural LMs by probing. A probe  \cite{alain:bengio:2016:probes,hupkes:2018:vis,conneau:2018:probing} is a diagnostic tool designed to extract linguistic representations encoded by another model. 

An ample body of research exists on probing  the sub-sentential structure of contextualized word embeddings. \citet{peters:etal:2018:dissecting} probed neural networks to see to what extent span representations capture phrasal syntax. \citet{tenney:2019:what} devised a set of \textit{edge probing} tasks to get new insights on what is encoded by contextualized word embeddings, focusing 
 on the relationship between spans rather than individual words. This enables them to go beyond sequence labeling problems to syntactic constituency, dependencies, entity labels, and semantic role labeling. Their results on syntactic constituency are in line with our findings. The major difference is that we employ simpler probes while achieving 
  similar results. Moreover, we separate the effect of semantics using corrupted data and we reconstruct full constituent trees using our probing setup. Most recently, ~\citet{perturbed:Masking:2020} propose a parameter-free probing technique to analyze LMs via perturbed masking. Their approach is based on accessing the impact that one word has on predicting another word within a sequence in the Masked Language Model task. They have also shown that LMs can capture syntactic information with their self-attention layers being capable of surprisingly effective learning. 

\citet{hewitt-manning-2019-structural} demonstrated, by using a structural probe, that it is possible to find a linear transformation of the space of the LM's activation vectors under which the distance between contextualized word vectors corresponds to the  distance between the respective words in the dependency tree. In a similar vein, \citet{chen2021probing} introduced another structural probe, \textit{Poincaré probe} and have shown that syntactic trees can be better reconstructed from the intermediate representation of BERT in a hyperbolic subspace.


\paragraph{Syntactic and semantic knowledge}
\citet{gulordava-etal-2018-colorless} and \citet{hall-maudslay-cotterell-2021-syntactic} recently argued that the work on probing syntax does not fully separate the effect of semantics while probing syntax. 
Both modify datasets such that the sentences become semantically nonsensical while remaining syntactically well-formed in order to assess, based on this data, whether a LM represents syntactic information. 
\citet{gulordava-etal-2018-colorless}
modify treebanks in four languages by replacing content words with other content words that have matching POS and morphological features.
They focus on the question if agreement information in the nonce sentences can be recovered from RNN language models trained on regular data.
\citet{hall-maudslay-cotterell-2021-syntactic} replaced words with pseudowords in a dependency treebank, and quantified how much the pseudowords affect the performance of syntactic dependency probes.
We followed a similar setup to separate out the effect of syntax from semantics. In contrast to \citet{hall-maudslay-cotterell-2021-syntactic}, we replace words with other words (not pseudowords) that occur in a similar syntactic context but are different semantically. 
This way, the LM has seen most or all words from the semantically nonsensical sentences in pretraining and has learned their syntactic properties.


\paragraph{Fine-grained LM analysis}
\citet{durrani-individual:emnlp20} used a unified diagnostic classifier approach to perform analyses at various granularity levels, extending  \newcite{dalvi:grain:2019}. We follow their approach and perform model-level, layer-level and neuron-level analyses. We additionally extend their approach by proposing an improved way to probe representations of two words. 


\section{Diagnostic Tasks and Constituency trees}\label{sec:diagnoseconst}

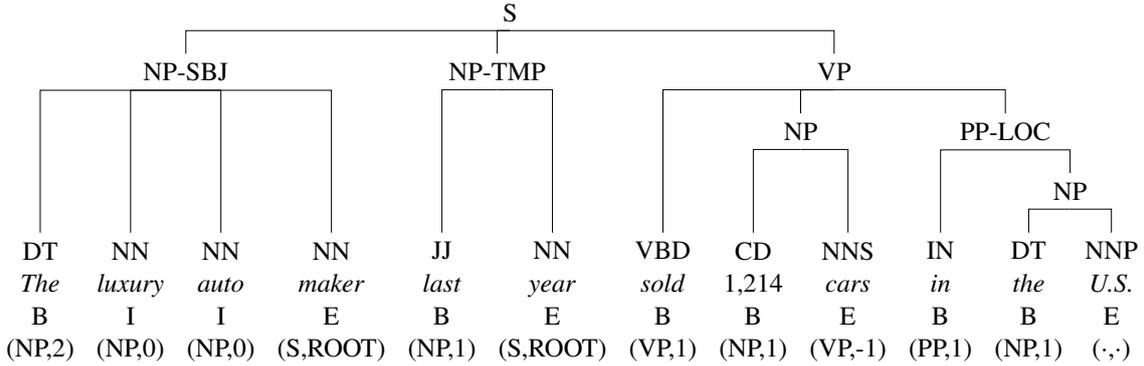
\begin{figure*}
    \centering
    
\begin{tikzpicture}
    \tikzset{level distance=25pt}
    \tikzset{every tree node/.style={align=center,anchor=north}}
    \tikzset{frontier/.style={distance from root=100pt}}
    \tikzset{edge from parent/.style= {draw,edge from parent path={(\tikzparentnode.south) -| (\tikzchildnode)}}}
    \begin{scope}[scale=0.9]
    \Tree[.S [.NP-SBJ DT\\\textit{The}\\{B}\\{(NP,2)} NN\\\textit{luxury}\\{I}\\{(NP,0)} NN\\\textit{auto}\\{I}\\{(NP,0)} NN\\\textit{maker}\\{E}\\{(S,ROOT)} ] [.NP-TMP JJ\\\textit{last}\\{B}\\{(NP,1)} NN\\\textit{year}\\{E}\\{(S,ROOT)} ] [.VP VBD\\\textit{sold}\\{B}\\{(VP,1)} [.NP CD\\1,214\\{B}\\{(NP,1)} NNS\\\textit{cars}\\{E}\\{(VP,-1)} ] [.PP-LOC IN\\\textit{in}\\{B}\\{(PP,1)} [.NP DT\\\textit{the}\\{B}\\{(NP,1)} NNP\\\textit{U.S.}\\{E}\\{($\cdot$,$\cdot$)} ] ] ] ]
\end{scope}
\end{tikzpicture}
    \caption{Example tree from the PTB. The line below the text shows gold labels for the simple chunking task. The bottom line shows label pairs from which the complete tree can be reconstructed.}
    \label{fig:ptb-example}
\end{figure*}

In this section, we define three classification tasks that are aimed at making different properties of constituency structure explicit. 
More specifically, the goal of these tasks is to make explicit if and how the LMs encode syntactic categories, such as \texttt{S}, \texttt{NP}, \texttt{VP}, \texttt{PP}. The first task, lowest common ancestor prediction, focuses on constituents that span large portions of the sentence. The second task, chunking, focuses on constituents with smaller spans. The third task focuses on complete syntactic trees. 

\noindent\textbf{Lowest common ancestor (LCA) prediction}
Let $s=w_0,\dots,w_n$ be a sentence.
Given combined representations of two tokens $w_i,w_j$ with $j \ge i$, predict the label of their lowest common ancestor in the constituency tree. LCA prediction is a multiclass classification task with 28 target classes (for the PTB).
In the example in Fig.~\ref{fig:ptb-example}, \textit{luxury} and \textit{maker} have the LCA \texttt{NP}: the lowest node dominating both words has label \texttt{NP} (ignoring the function tag \texttt{SBJ}).
The task also predicts LCA of two identical tokens. 
In this case, the lowest phrasal node above the token is the target label (for example, \texttt{VP} is the target label for \textit{sold}).

\noindent\textbf{Chunking}
For each token $w_i$, predict whether it is the beginning of a phrase (\texttt{B}), inside a phrase (\texttt{I}), the end of a phrase (\texttt{E}) or if the token constitutes a single-token phrase (\texttt{S}). A token can be part of more than one phrase, and in this case we consider the shortest possible phrase only. For example, $1,214$ in Fig.~\ref{fig:ptb-example} has label \texttt{B} because it marks the beginning of a noun phrase.
We also propose a version of this task with finer labels that combine \texttt{B,I,E,S} with the different phrase labels. In the detailed tagset, $1,214$ receives the label \texttt{B-NP}.

\noindent\textbf{Reconstructing full constituent trees}
\citet{vilares2020parsing} considered constituency parsing as a multi-label sequence labeling problem. For each token $w_i$, three labels are predicted: First, the label of the LCA of the token pair $(w_i,w_{i+1})$. Second, the depth of the LCA of $(w_i,w_{i+1})$ in the tree, relative to the depth of $(w_{i-1}, w_i)$.
Third, if the first token is a single-word constituent, and the label of the internal tree node directly above $w_i$. (Tokens in multiword constituents make up $>90\%$ of the data and receive a negative label.).
For the first two classifications, see Fig.~\ref{fig:ptb-example}. We build separate linear classifiers for each of these tasks and use their predictions to reconstruct full constituent trees. 

\section{Methods}
\label{sec:methods}

\noindent\textbf{Diagnostic classification}
A common method to reveal linguistic representations learned in contextualised embeddings is to train a classifier, a probe, using the activations of the trained LM as features. The classifier performance  provides insights into the strength of the linguistic property encoded in contextualised word representations. For all our experimental setups, we employ the NeuroX toolkit \cite{dalvi:2019:neurox} for diagnostic classification, as it confers several mechanisms to probe neural models on the level of model, layers and neurons.

\noindent\textbf{Layer-level probing} 
We probe the activations of individual layers with linear classifiers to measure the linear separability of the syntactic categories at layer-level. The performance at each layer serves as a proxy to how much information it encodes with respect to a given syntactic property.

\noindent\textbf{Neuron-level probing}  Layer-wise probing 
cannot account for all syntactic abstractions encoded by individual neurons in deep networks. Some groups of neurons that are spread across many layers might robustly respond to a given linguistic property without being exclusively specialized for its detection. By operating also 
 at the level of the neurons, we aim at separating the most salient neurons across the network that learn a given linguistic property. We conduct a \textit{linguistic correlation analysis}, as proposed by \citet{dalvi:grain:2019}.
 The linear classifier is augmented 
 with elastic net regularization \cite{regularization:2005}, which strikes a balance between selecting very focused features (neurons) and distributed features across many properties. 
The input neurons to the linear classifier are ranked by saliency with respect to the classification task. 



\noindent\textbf{Input Representation}
 We combine the representation vectors $x_i,x_j \in \mathbb{R}^r$ of two tokens in LCA prediction and parse tree reconstruction via concatenation ($concat(x_i,x_j)\in \mathbb{R}^{2r}$).
In all experiments, $concat$ produced significantly better results than elementwise averaging or a variant of the maximum. We cover the latter methods in Apps.~\ref{app:tok:comb} and \ref{app:lca:res}.


\section{Experimental Setup}\label{sec:expsetup}
\subsection{Data\label{sec:data}}

We use data from the English Penn Treebank \cite[PTB, ][]{marcus-etal-1993-building} for all our experiments. As preprocessing, 
 we remove punctuation and null elements 
  from the trees.   
The original dataset makes use of fine-grained category labels that consist of the syntactic category and function tags. 
Function tags indicate grammatical (such as \texttt{SBJ} for subject) or adverbial (\texttt{LOC} for locative) information. 

\subsubsection{Sampling}
\label{sec:sec:sampling}



\noindent\textbf{LCA}
For LCA prediction, we remove all function tags to keep the number of target labels small. We train on 100k randomly selected token pairs from all sentences in the PTB training sections. For evaluation, we use the first 200 sentences of the PTB development set with a sentence length $\le 20$. 
We train on a sampled subset of the training split of the PTB because the distribution of target labels is highly imbalanced for full sentences: 
Most of the token pairs have a relatively large distance in the constituent tree, and their LCA is a node very high in the tree (typically with a label for some kind of sentence, such as \texttt{S} or \texttt{SBAR}). In addition, some phrase labels are less frequent than others (see App.~\ref{app:label:distr}). 
Since we are interested in a probe that performs well across target labels, we randomly sample token pairs from all sentences in the training data such that for each label, the relative frequency of the label in the sampled data is the mean between the original relative frequency in the data and a uniform distribution. Consider a label $y\in Y$, where 
$f(\texttt{y})$ 
denotes the relative frequency when considering all token pairs. The relative frequency $f_s$ in the sampled training data is defined as:
\begin{align}
    f_s(y) = (f(y) + \frac{1}{|Y|})*0.5
\end{align}
This 
ensures that all category labels in the PTB are adequately represented in the training data of the LCA prediction, and that the training data 
covers a diverse set of linguistic phenomena. 

\paragraph{Chunking}
We train on the first 10k sentences of the training split and evaluate on the first 2,5k sentences of the development split. The label distribution is relatively balanced and requires no sampling.


\begin{table*}									
\centering
\small
\begin{tabular}{l|l}									
\toprule									  
orig. & Pierre Vinken, 61 years old, will join the board as a nonexecutive director Nov. 29. \\	
.33  & Pierre \textbf{Berry}, \textbf{5,400} years old, \textbf{shall} join the board as a nonexecutive director Nov. 29.\\
.67 & \textbf{Mesnil Vitulli}, \textbf{9.76 beers state-owned}, \textbf{ca succeed either} board as a \textbf{cash-rich} director \textbf{October} \textbf{213,000}.\\\midrule
orig. & Mr. Vinken is chairman of Elsevier N.V., the Dutch publishing group.\\
.33 & Mr. Vinken is \textbf{growth} \textbf{without} Elsevier \textbf{Hills}, \textbf{each} Dutch publishing group.\\
.67 & \textbf{Tata Helpern s} chairman \textbf{plus} Elsevier \textbf{Ohls}, \textbf{a} Dutch \textbf{snaking} group.\\
   \bottomrule
    \end{tabular}
    \vspace{-2mm}
    \caption{Examples for the manipulated data. Replaced words are printed in boldface. }
    \vspace{-2mm}
\label{tab:examples-manipulated-data}						    
\end{table*}

\subsubsection{Syntactic and semantic knowledge}\label{sec:data:syn:sem}
To ensure that the probing classifier captures syntactic properties and not semantic properties, we use the original PTB data as well as two modified versions of the PTB with semantically nonsensical sentences that have the same syntactic structure as the original data. 
The modified versions of the PTB are obtained by making use of the dependency PTB \citep{de-marneffe-etal-2006-generating}:
\begin{enumerate}
    \item Record the dependency context of each token in the dataset. The dependency context consists of (i) the POS tag, (ii) the dependency relation of the token to its head, and (iii) the list of dependency relations of the token to its dependents. \vspace{-.5em}
    \item Replace a fraction of tokens with other tokens that also appear in the dataset in the same dependency context.
\end{enumerate}
Two versions are created, replacing either a third or two thirds of the tokens. See 
 Table \ref{tab:examples-manipulated-data} for two examples. 
When creating manipulated datasets, we separate the training and evaluation data. 
To create manipulated training data, 
we look for token replacements in the training split of the PTB (PTB sections 0-18). 
For manipulated evaluation data, we look for token replacements in the development and test splits of the PTB (sections 19-24). 
This ensures that the training and evaluation data do not mix, and at the same time, meanings of the newly created sentences are as diverse as possible.

\subsection{Probing Classifier Settings}
We use linear classifiers trained for 10 epochs with Adam optimization, an initial learning rate of $0.001$ and elastic net regularization parameters $\lambda_1=\lambda_2=0.001$.
The representation for an input  token 
 is created by averaging the representations of its subword tokens in the LM tokenizer.

\subsection{Transformer Models}
We train structural probes on three 12-layered pretrained transformer models; the base versions of BERT \cite[uncased,][]{devlin-etal-2019-bert}, XLNet \cite[cased,][]{xlnet-NEURIPS2019_dc6a7e65}, and  RoBERTa \cite{liu2019roberta}, and a 6-layered model; DistilBERT \cite[uncased,][]{sanh2020distilbert}. This provides an opportunity to compare the syntactic knowledge learned in models with 
different hyperparameter settings and pretraining objectives.




\subsection{Baselines}

We use three baselines to put the results of our probes into context. 

\noindent\textbf{Random BERT} The first baseline is used in all experiments. It evaluates how much information about linguistic context is accumulated in the LM during pretraining \cite{belinkov:2021:probing}. The model for this baseline has the same neural architecture, vocabulary and (static) input embeddings as BERT base, but all transformer weights are randomized. 

\noindent\textbf{Selectivity} To evaluate if the probe makes linguistic information explicit or just memorizes the tasks, we use the control task proposed by \newcite{hewitt-liang-2019-designing} and described in App.~\ref{app:controltask}. The difference between control task performance and linguistic task performance is called selectivity. 
The higher the selectivity, the more one can be sure that the classifier makes linguistic structure inside the representations explicit and does not memorize the task.
This baseline is used for the chunking and, in a modified version, the LCA experiments.

\noindent\textbf{Individual tokens} This baseline evaluates how much the representation of each token, in contrast to token pairs, contributes to the overall performance. This baseline is used only for the LCA experiments. 
We train two classifiers using the representation of either the first token in the pair or the second token and evaluate the performance on the diagnostic tasks.  
The trees in the PTB are right-branching. The left token in most cases is closer to the LCA node than the right token, thus we expect that the classifier trained on only the left token has a better overall performance.


\section{Results for LCA prediction and chunking}\label{sec:results}

We experimented using four pretrained models. Due to the limited space and the consistency of results, we reported the analysis for the RoBERTa model only in most of the cases. The complete results of all models are shown in appendix sections \ref{app:chunk:res} and \ref{app:lca:res}.
%
In the following, we 
 assess 
  the overall performance of the probing classifiers on both linguistic tasks.  
  Then, we evaluate how changing the semantic structure of the data influences the probing classifiers. Lastly, we show some insights into layer-level and neuron-level experiments.


\subsection{Overall performance}
Tab.~\ref{fig:results} shows the performance of the classifiers 
 trained on non-manipulated data using all neurons of the network (orig./orig.). We observed high performance for each diagnostic task. 
The differences to the 
 baselines show that the knowledge about the task is indeed learned in the representation. 

\begin{table} 
    \centering
    \small
\begin{tabular}{l|l|lll}
\toprule
\multirow{2}{*}{} & \multirow{2}{*}{train/test} & \multicolumn{3}{c}{RoBERTa}\\
& & task & sel. & $\Delta_{Random}$ \\\midrule
\multirow[c]{5}{*}{\parbox{1.5cm}{LCA\\ $concat$}} & orig./orig. &  \textbf{84.4} & 59.9 & 26.5 \\
     & $.33$/orig. & 81.0 & \textbf{62.1} & 25.4 \\
     & $.33$/$.33$ & 78.9 & 56.0 & 26.2 \\
     & $.67$/orig. & 78.6 & 60.7 & \textbf{29.5} \\
     & $.67$/$.67$ & 73.7 & 56.2 & 25.2 \\\midrule
\multirow[c]{5}{*}{\parbox{1.5cm}{chunking\\ simple}} & orig./orig. & \textbf{95.9} & 16.9 & \textbf{27.3}\\
     & $.33$/orig. & 94.7 & 16.2 & 25.2 \\
     & $.33$/$.33$ & 92.7 & 18.7 & 25.6 \\
     & $.67$/orig. & 93.6 & 19.4 & 22.2 \\
     & $.67$/$.67$ & 90.2 & \textbf{20.5} & 21.2 \\\midrule
\multirow[c]{5}{*}{\parbox{1.5cm}{chunking\\ detailed}} & orig./orig.& \textbf{90.7} & 14.2 & \textbf{38.7} \\
     & $.33$/orig. &  88.3 & 12.1 & 31.8 \\
     & $.33$/$.33/$ & 86.1 & \textbf{15.3} & 33.4 \\
     & $.67$/orig. &  85.3 & 12.3 & 30.1 \\
     & $.67$/$.67$ &  81.5 & 14.3 & 31.3 \\\bottomrule
\end{tabular}
\vspace{-2mm}
\caption{Results on different datasets. For each task, there are different setups where the model is trained and evaluated on the unchanged treebank (orig.), and two versions with either a third (0.33) or two thirds (0.67) of the tokens replaced.
'task' shows the performance on test set. 'sel.' shows the selectivity (difference to control task), and $\Delta_{Random}$ shows the performance difference to the Random BERT model.
}
\vspace{-2mm}
\label{fig:results}
\end{table}

\noindent\textbf{LCA prediction}
The best results are achieved when concatenating token representations (84.4\% acc.). 
For other representation methods, see App.~\ref{app:lca:res}.
We additionally consider single word representations from the word pair as input. The left token representations are better predictors (66.5\% acc.~on orig./orig.) than those from the right token (40.8\%). 
The large differences between $concat$ and the baselines shows that 
the probe is not memorizing the task, and that information relevant for predicting LCA is acquired during pretraining.


\noindent\textbf{Chunking}
Chunking detailed (90.7\% acc.) is a harder task than chunking simple (95.9\%). Although the classifier for the detailed tagset shows relatively low selectivity in comparison to chunking simple, the overall selectivity is high enough to claim that the knowledge about these probing tasks is learned in the representation. 
The difference to the Random BERT model is higher for chunking detailed than for chunking simple, which shows that fine-grained syntactic knowledge is indeed learnt during pretraining.

\subsection{Does the probe learn syntax or semantics?}

The high performance of the classifiers serves as a proxy to the amount of syntactic knowledge learned in the representations. 
However, due to the presence of semantic cues in the data, high performance of a syntactic probe may not truly reflect the learning of syntax in the model. To investigate this, we manipulated our diagnostic task data (Sec. \ref{sec:data:syn:sem}) to separate syntax from semantics, and 
 then trained the probing classifiers on the manipulated data. 

The second column in Table \ref{fig:results} shows variations of the manipulated data. The classification performance dropped slightly on the diagnostic tasks at 0.33/orig. Moreover, the classifier performed slightly better when evaluating on original data ($*$/orig.) compared to 
 manipulated data (such as .33/.33). There are two possible reasons for this: First, the probing classifiers may still rely on semantic knowledge, even when  trained on the manipulated data; Second, it is possible that the manipulated data contains syntactically ill-formed sentences. 
 Nonetheless, performance and differences to baselines are reasonably high and 
give good reason to believe that the classifiers  are able to
 extrapolate syntactic knowledge even from semantically nonsensical data. 
We now proceed with summarizing 
 what our experiments tell about syntactic knowledge in specific layers/neurons of the LM.

\begin{figure}
    \centering
    \includegraphics[width=.49\linewidth]{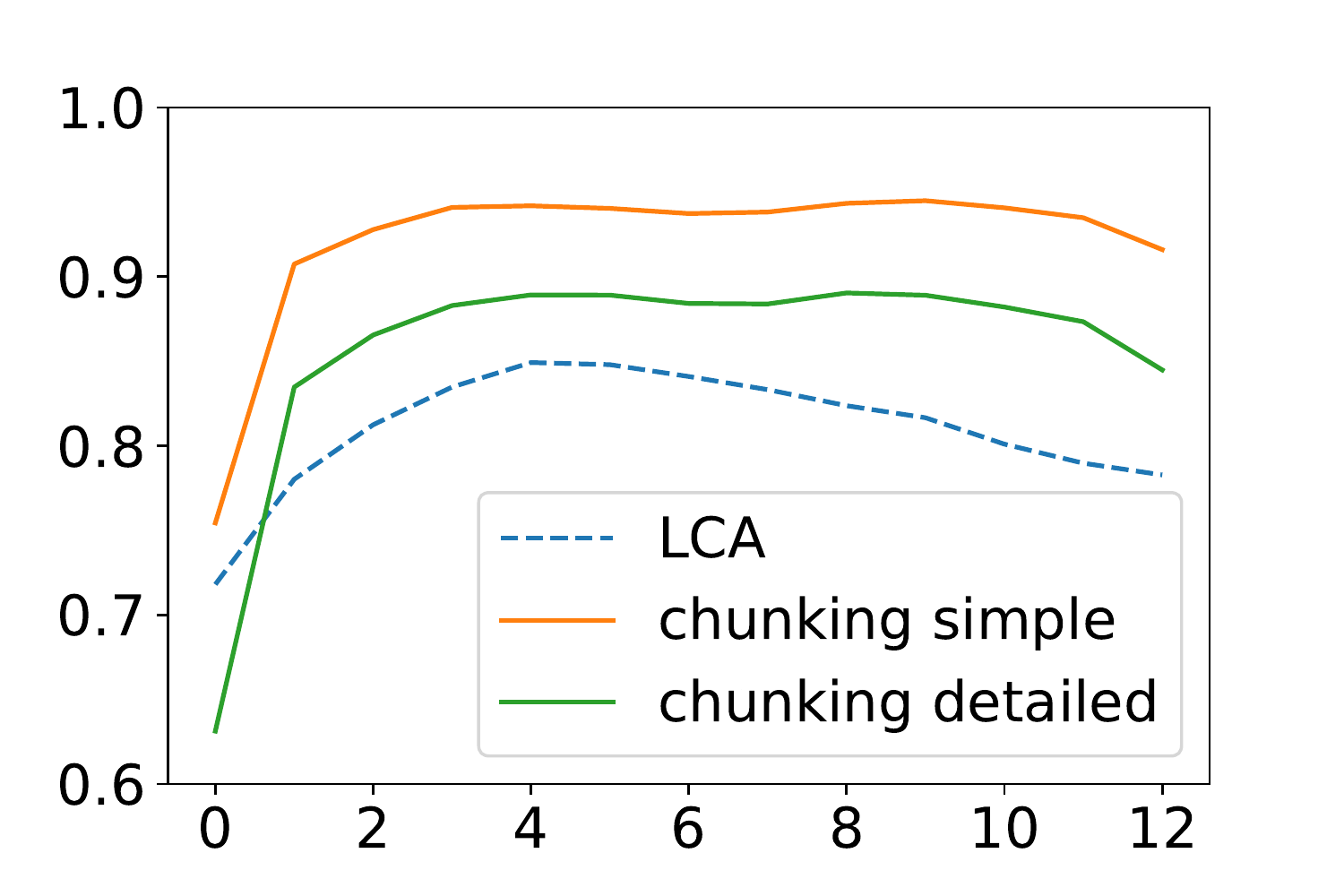}
    \includegraphics[width=.49\linewidth]{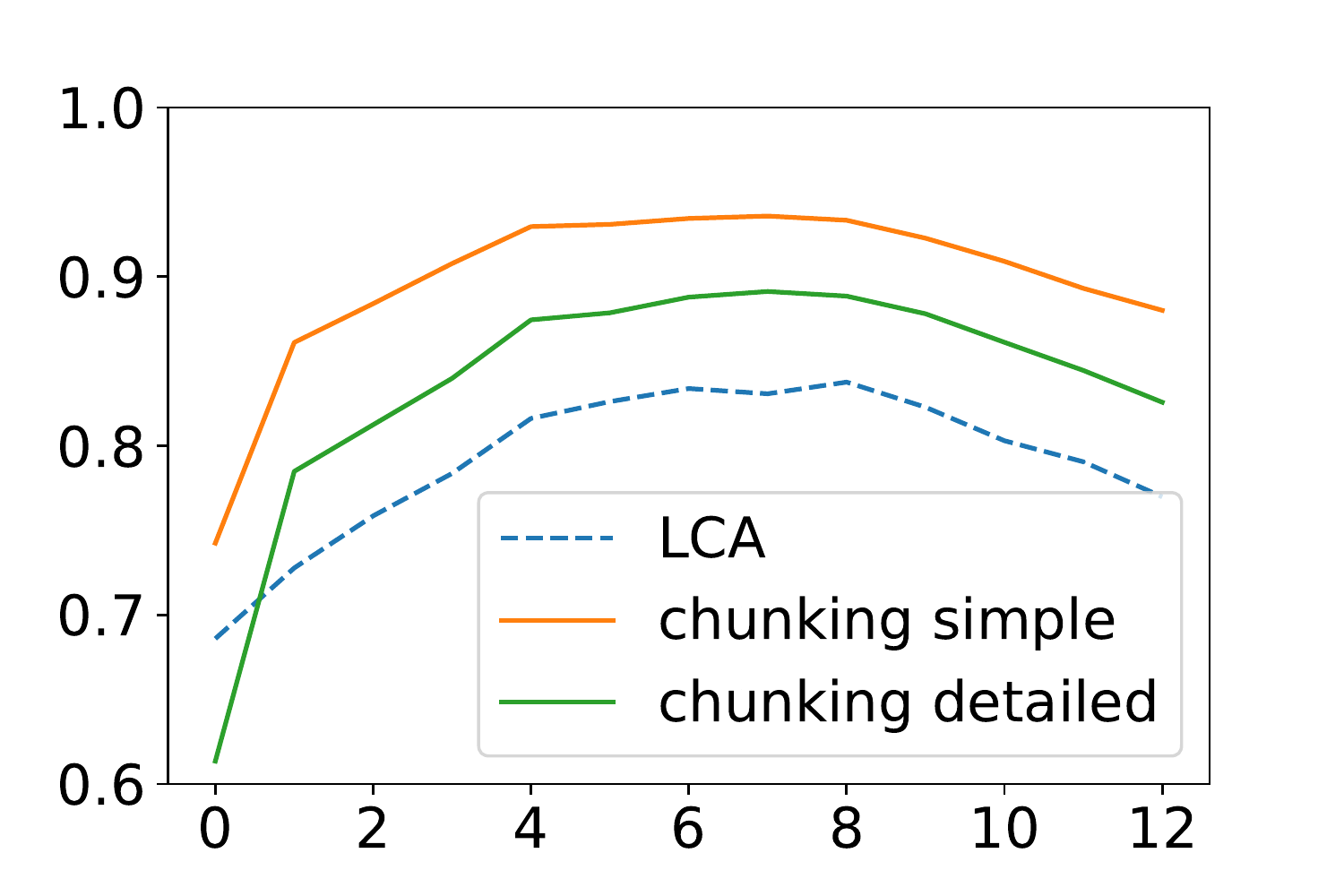}
    \caption{Acc. of layer-wise retraining 
     probing classifiers on RoBERTa (left) and BERT (right) representations for non-manipulated data.}
    \label{fig:layerwise-all-tasks}
\end{figure}

\begin{figure}[t]
    \centering
    \includegraphics[width=.49\linewidth]{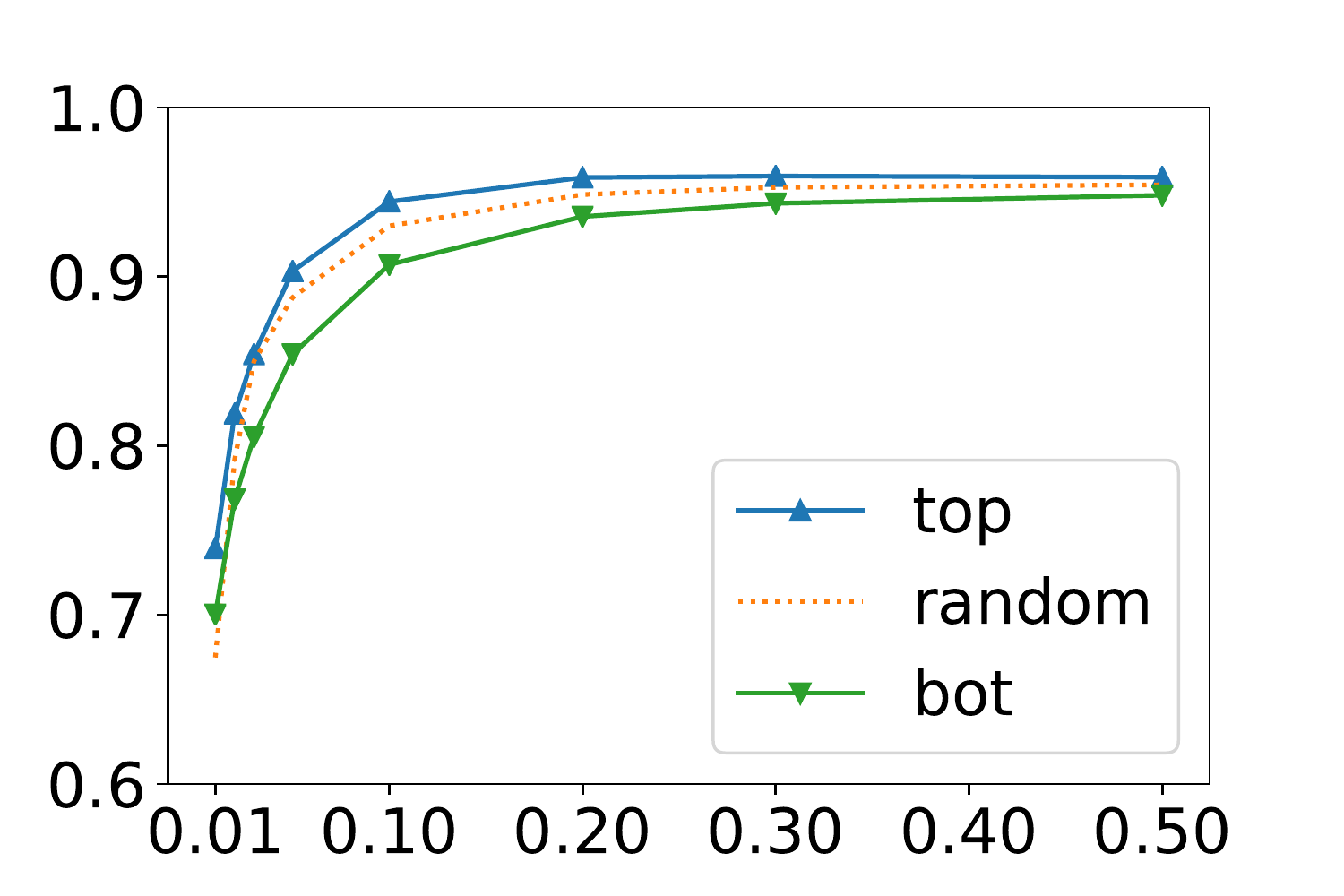}
    \includegraphics[width=.49\linewidth]{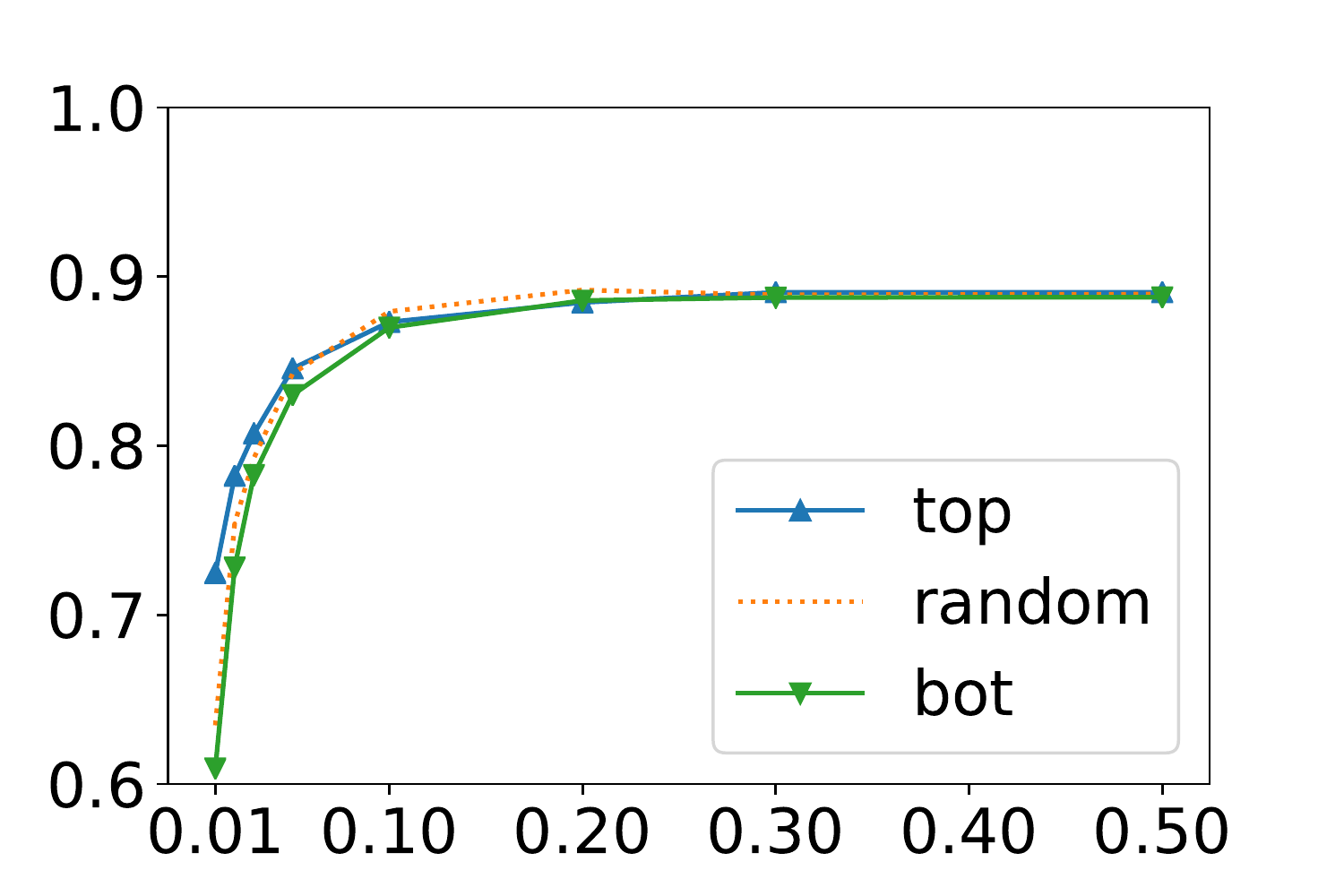}
    \caption{Acc. for simple (left) and detailed (right) chunking tagsets, on top, random and bottom neurons for RoBERTa on non-manipulated data. The horizontal axis plots the fraction of neurons that is selected from all layers.}
    \label{fig:acc-min-neurons}
\end{figure}

\subsection{Layer-wise results}\label{sec:layer-wise-results}

        

Fig.~\ref{fig:layerwise-all-tasks} shows how syntactic knowledge is distributed across all layers. The embedding layer performed worse while middle layers showed the best results, i.e., syntactic information is better represented in the middle layers. The highest layers are more heavily influenced by the pretraining objective, 
  which explains the consistent performance drop across models and tasks on the last layers.

Comparing layer-wise performance with the overall performance, none of the individual layers outperformed the classifier trained on 
 all layers 
  for chunking. In the case of LCA prediction, the performance of layers 4-5 in RoBERTa (6-8 in BERT) are better than the overall performance on 
   all layers. 
   Comparing models, we observed that RoBERTa learns the syntactic knowledge much earlier in the network compared to BERT (see the relatively sharp rise of performance in the lower layers of RoBERTa).  


\subsection{Neuron-level results}

In this section we carry out a more fine-grained neuron-level analysis of the
representations. Linguistic correlation analysis~\cite{dalvi:grain:2019} provides a ranking of neurons with respect to the diagnostic task. 

\noindent\textbf{Minimum Subset of Neurons}
We evaluated the neuron ranking by training  classifiers using top/bottom/random N\% neurons. Fig.~\ref{fig:acc-min-neurons} shows the accuracy curves for the chunking task. 
The performance margin between different selected neurons is very low. This shows that syntactic information can be extracted from any relatively small subset of neurons i.e. $20-30\%$ of neurons suffice for a probing classifier to perform with the same accuracy as when trained on full representations.
Neuron ranking on combined representations does not work well: In some cases, performance on a fraction of randomly selected neurons is worse than performance on the same fraction of neurons ranked as important (see App.~\ref{app:neuron:lca}).

\begin{figure}[tb]
    \centering
    \begin{subfigure}{.9\linewidth}
        \includegraphics[width=.9\linewidth]{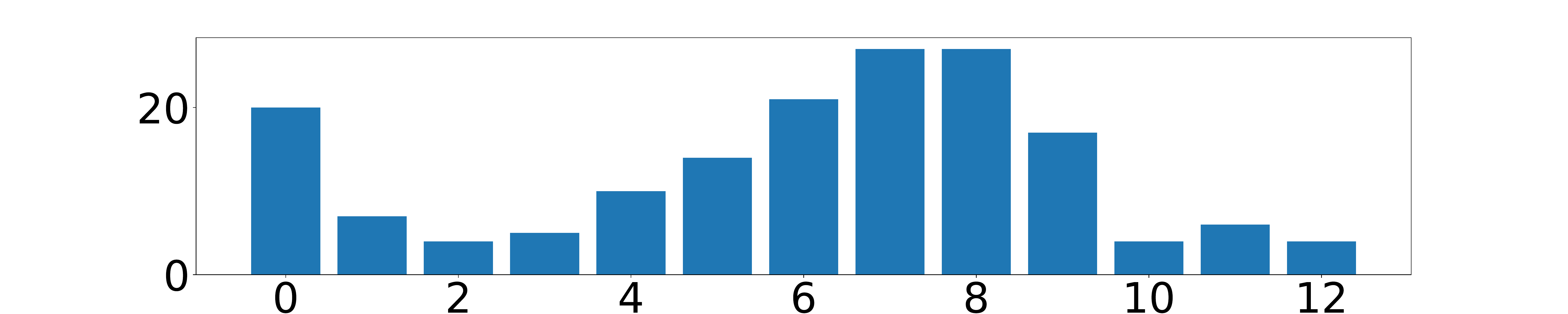}
        \caption{BERT}
    \end{subfigure}
    
    \begin{subfigure}{.9\linewidth}
        \includegraphics[width=.9\linewidth]{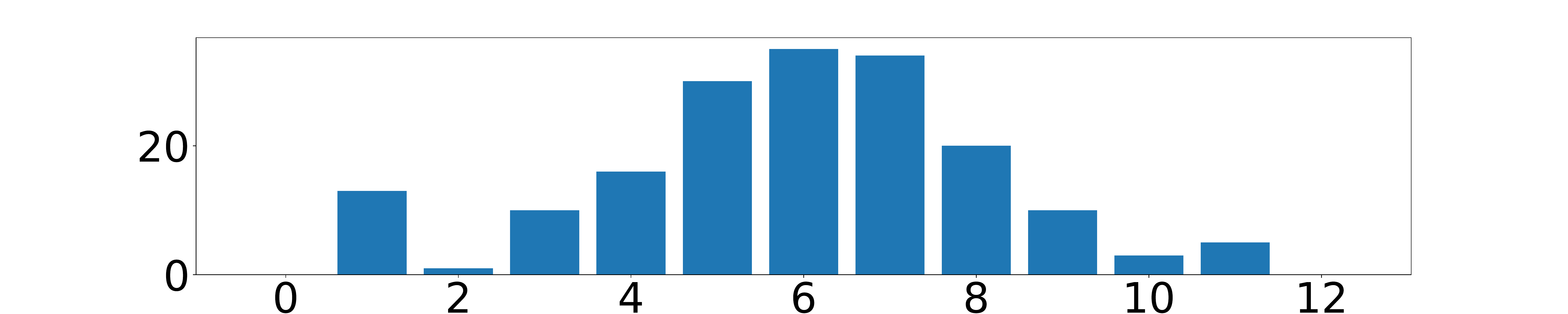}
        \caption{XLNet}
    \end{subfigure}
    
    \begin{subfigure}{.9\linewidth}
        \includegraphics[width=.9\linewidth]{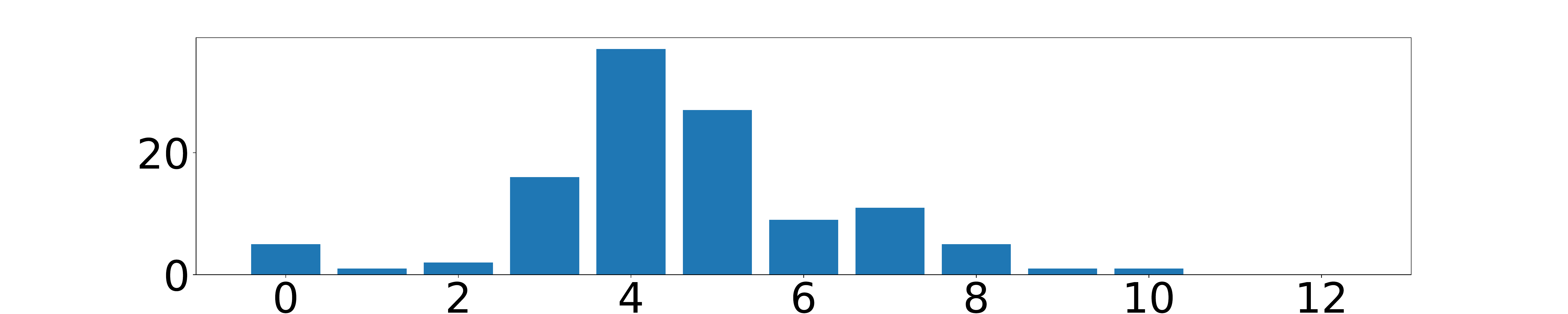}
        \caption{RoBERTa}
    \end{subfigure}
    \vspace{-2mm}
    \caption{Spread of neurons relevant for recognizing \texttt{S} in LCA prediction, across layers.}
    \vspace{-2mm}
    \label{fig:spread-per-prop-s}
\end{figure}

\begin{figure}[t]
    \centering
    \includegraphics[width=.9\linewidth]{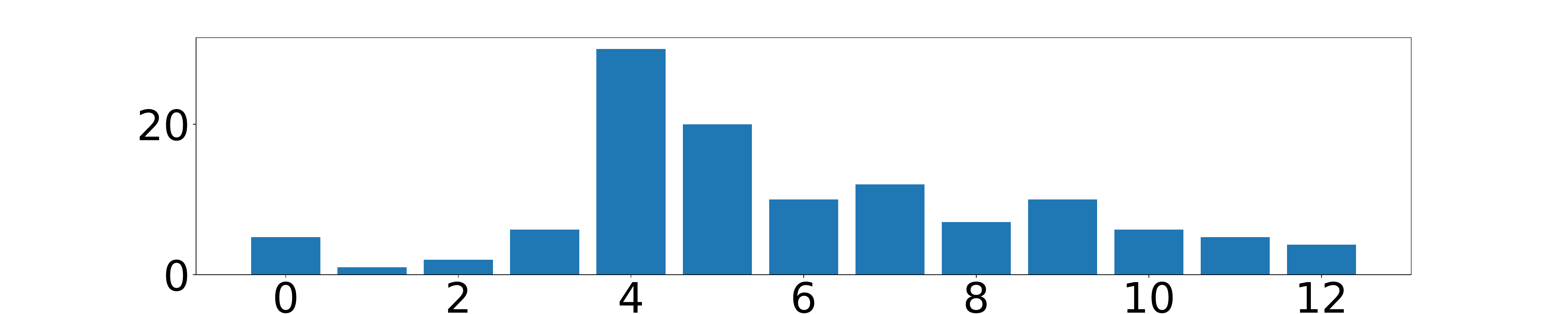}
    \vspace{-2mm}
    \caption{Spread of neurons in RoBERTa across layers that are relevant for identifying \texttt{NP}s in LCA prediction}
    \vspace{-2mm}
    \label{fig:spread-per-prop-np}
\end{figure}

\noindent\textbf{Distribution of Neurons for LCA prediction}
Training on subsets of the neurons for LCA prediction is problematic, because the neuron ranking list contains neurons from both token representations.
Even though, the distribution of salient neurons across layers yields interesting insights.
Fig.~\ref{fig:spread-per-prop-s} presents the spread of top selected neurons for \texttt{S}. 
As in Sec.~\ref{sec:layer-wise-results}, we found again 
 that top neurons learning syntactic properties come 
  from the middle layers. 
For 12-layer LMs, we see a trend that neurons from the positional encoding in the embedding layer are utilized to identify distant tokens with LCA \texttt{S}.
When comparing 
the salient neurons 
selected from 
each layer,
we observe that for identifying \texttt{S}, neurons from the highest layers are less relevant than when identifying \texttt{NP}s (Fig.~\ref{fig:spread-per-prop-np}). This might be due to the comparatively high structural diversity we find in \texttt{NP}s.\footnote{All models are more accurate in LCA prediction when the two tokens are more distant, see App.~\ref{app:lca:res}. Large distance between tokens correlates with LCA nodes close to the root of the syntactic tree, where the LCA often has label \texttt{S}.}

\noindent\textbf{Neurons learning Syntax vs. Semantics}
Comparing the neuron rankings of chunking classifiers trained on the different datasets shows that there is relatively little overlap between the different groups of highest-ranking neurons (see App.~\ref{fig:overlap-datasets}). 
This means that the probing classifiers focus 
 on different neurons when training on manipulated data, compared to 
  the original data. 
Presumably, the probe focuses more on syntactic and less on semantic information when trained on manipulated data.

\section{Reconstructing full Parse Trees}\label{sec:tree-reconstruction}

With the insights gained in the previous section,
we test if full constituency trees can be linearly separated from LM representations. 
For this, we train 3 linear classifiers that take as input the concatenated representations of two adjacent tokens and predict the three labels described in Sec.~\ref{sec:diagnoseconst}.
The classifiers for this task take as input not the full LM representation from all layers, but instead the concatenations of every third layer for the 12-layer LMs, and every second layer for the 6-layer LM. 
This way, the input dimensionality of the classifier is restricted, but the probe can use information from different parts of the LM.
The probe is trained (evaluated) on all 38k (5.5k) sentences of the training (development) split of the PTB. We find that the constituency trees reconstructed from different LMs are of high quality (Tab.~\ref{tab:full-trees-results}, App.~\ref{app:results:parsetrees}). 
We achieve a labeled F1 score of 82.6 on the non-manipulated dataset for RoBERTa (80.5 for XLNet, 80.4 for BERT) which is 31 points better than the Random BERT baseline. 
This outperforms the result from \newcite{vilares2020parsing} for BERT by 2.2 points. They also use a linear classifier, but their classifier receives as input only the final layer representation of BERT for the first token in the token pair.  

When comparing trees reconstructed from different LMs against each other, we find however that they are quite different. For example, comparing the sentence-level F scores for trees reconstructed from XLNet to 
 those from RoBERTa yields a Pearson correlation of 0.52 only 
 (compared to 0.64 for DistilBERT and BERT, see App.~\ref{app:comp:parsetrees} for the full comparison).
This shows that different syntactic properties are linearly separable from the representations of different LMs. 
These results are not a shortcoming of our probe. We reconstructed parse trees from RoBERTa for the same dataset twice, and a comparison of the two sets of trees gave 
 a labeled F1 score of 96.3. 
We conclude from this that LMs trained on different data and towards different objectives, such as RoBERTa and XLNet, implicitly make different syntactic generalizations. This insight might have implications for parsing (which is not in the scope of our paper): combining embeddings from both LMs might improve parsing results, compared to using just 
 one LM, as usually done. 

\begin{table}[t]
    \centering
    \small
\begin{tabular}{l|rr}
\toprule
{} &  RoBERTa &  $\Delta_{Random}$ \\
\midrule
orig.     &         \textbf{82.6} &   31.2\\
.33/orig. &         80.9 & 31.3\\
.33/.33  &         77.8 & \textbf{31.7}\\
.67/orig. &         78.3 &  30.4\\
.67/.67  &         72.8 &  28.1\\
\bottomrule
\end{tabular}
\vspace{-2mm}
    \caption{Labeled F1 scores for all datasets for reconstructing full parse trees}
    \vspace{-2mm}
    \label{tab:full-trees-results}
\end{table}

\section{Conclusions}\label{sec:conclusion}

Our experiments have shown that different pre-trained LMs 
 encode fine-grained linguistic information that is also present in constituency trees.
More specifically, LMs 
 are able to identify properties of different types of constituents, such as S, NP, VP, PP. 
Good results on the chunking task show that the classifiers are able to combine knowledge about the kind of constituents that a token is part of, and knowledge of the position of a token in the constituent. 
Using the sequence labeling tasks presented in \newcite{vilares2020parsing}, we have shown that full constituency trees are linearly separable from four different pretrained LMs with high quality - even for semantically nonsensical data.
In line with \citet{gulordava-etal-2018-colorless}, 
we observe a moderate performance drop between performance on the original and nonce dataset. 
The performance drop is smaller than in \citet{hall-maudslay-cotterell-2021-syntactic}.
who use English pseudowords which the LM has probably never encountered. We use English words whose syntactic and semantic properties are already well-established inside the LM.
In future work, we plan to extend this syntactic probing approach to other languages and other syntactic annotation schemes \citep[for instance][]{hockenmaier-steedman-2007-ccgbank,evang-etal-2021-bootstrapping}.






\bibliography{biblio}
\bibliographystyle{acl_natbib}

\appendix

\onecolumn

\section{Appendix}

\subsection{Constituency and Dependency Trees}\label{app:const:vs:dep}

Here, you see the constituency tree (above the sentence) and the dependency tree (below) for a simple sentence, following the annotation scheme of the PTB and its dependency version. More fine-grained hierarchical structure is encoded in the node labels of the constituency structure. For example, the constituency tree assigns the label VP to the spans \textit{walking on the moon} and \textit{am walking on the moon}. However, these spans are not reflected by particular entities in the dependency tree.

\begin{center}
    
\begin{tikzpicture}[sibling distance=32pt]
\tikzset{level distance=25pt}
\tikzset{every tree node/.style={align=center,anchor=north}}
\tikzset{frontier/.style={distance from root=140pt}}
\tikzset{edge from parent/.style= {draw,edge from parent path={(\tikzparentnode.south) -| (\tikzchildnode)}}}

\Tree[.S 
    [.NP-SBJ \node (eye) {PRP\\\textit{I}}; ]
    [.VP \node (am) {VBP\\\textit{am}};
    [.VP \node (walking) {VBG\\\textit{walking}};
    [.PP-LOC \node (on) {IN\\\textit{on}}; [.NP \node (thee) {DT\\\textit{the}}; \node (moon) {NN\\\textit{moon}}; ] ]
    ]
    ] ]

\node (root) at (70pt,-200pt) {ROOT};
\draw[->,in=-90,out=-150] (walking) edge node[anchor=north] {aux} (am);
\draw[->,in=-80,out=-120] (walking) edge node[anchor=north] {nsubj} (eye);
\draw[->,in=-90,out=-60] (walking) edge node[anchor=north] {dobj} (moon);
\draw[->,in=-90,out=-150] (moon) edge node[anchor=north] {det} (thee);
\draw[->,in=-80,out=-120] (moon) edge node[anchor=north] {case} (on);
\draw[->] (root) -- (walking);
\end{tikzpicture}
\end{center}

\subsection{Control tasks and Selectivity}\label{app:controltask} To ascertain that our structural probe learns complex structural generalization and does not memorize structural information from the task, following \citet{hewitt-liang-2019-designing}, we design two control tasks (CT). This consists in randomizing the labels of syntactic categories, creating a new dataset. In our implementation, the distribution of randomly selected labels approximates the class distribution in the training set.
For chunking, a numerical target label is assigned randomly to each word type. 
We slightly modify this baseline to be able to handle token pairs. For LCA prediction, we assign a random numeric label to each pair of word types. 
For example, the word pair (\textit{the}, \textit{sold}) in Fig.~\ref{fig:ptb-example} always receives the label $1$, regardless of the context where it occurs. And since the dataset contains more word type pairs than individual word types, the control task setup is inherently more complex for LCA prediction than the one for chunking.

\subsection{Token combination methods}\label{app:tok:comb}

For LCA prediction, w conducted experiments using three different combination methods:
(i) concatenation ($concat(x_i,x_j) \in \mathbb{R}^{2r}$, see Sec.\ref{sec:methods});
(ii) element-wise average $avg(x_i,x_j) \in \mathbb{R}^{r} $; 
and (iii) element-wise signed absolute maximum \maxs of two scalars $m,n$. $max_s(w_i,w_j)\in \mathbb{R}^r$ prefers strong positive and negative neuron activations while keeping the sign of the activation value in the combined vector:
\begin{align}
    max_s(m,n) = \begin{cases}
    m &\text{if } |m| > |n| \\
    n & \text{otherwise}
    \end{cases} 
\end{align}
Averaging is 
 the most lossy combination: 
  Large positive or negative neuron activations are canceled out if they are not shared between both vectors. 
 Concatenation is the only lossless combination. 
  The concatenation result 
 $concat(x_i,x_j) \in \mathbb{R}^{2r}$ has a larger dimensionality than the other combination methods. 

\subsection{Label distributions for the different probing tasks}\label{app:label:distr}
\begin{table}[h]

\begin{subtable}[b]{.5\textwidth}
    \centering\footnotesize
\begin{tabular}{l|rrrr}
\toprule
{}
&      B &     I &      E &     S \\
\midrule
VP       &  13.77\% &  1.24\% &   0.07\% &  0.96\% \\
NP       &  12.53\% &  8.83\% &  12.27\% &  4.05\% \\
PP       &   9.73\% &  0.25\% &   0.00\% &  0.03\% \\
NP-SBJ   &   2.96\% &  1.73\% &   3.11\% &  2.46\% \\
SBAR     &   1.14\% &  0.10\% &   0.00\% &     \\
ADJP     &   0.82\% &  0.21\% &   0.50\% &  0.38\% \\
QP       &   0.51\% &  0.75\% &   0.85\% &  0.00\% \\
ADVP     &   0.35\% &  0.04\% &   0.27\% &  1.57\% \\
NP-TMP   &   0.28\% &  0.07\% &   0.29\% &  0.14\% \\
S        &   0.25\% &  0.29\% &      &     \\
NP-PRD   &   0.18\% &  0.18\% &   0.18\% &  0.03\% \\
NP-ADV   &   0.16\% &  0.01\% &   0.16\% &  0.01\% \\
NP-LGS   &   0.12\% &  0.16\% &   0.13\% &  0.02\% \\
NP-EXT   &   0.08\% &  0.01\% &   0.03\% &  0.03\% \\
NX       &   0.06\% &  0.06\% &   0.06\% &  0.03\% \\
NAC      &   0.05\% &  0.04\% &   0.00\% &  0.00\% \\
WHPP     &   0.04\% &     &      &     \\
CONJP    &   0.04\% &  0.02\% &   0.04\% &  0.00\% \\
WHNP     &   0.03\% &  0.01\% &   0.04\% &  0.75\% \\
UCP      &   0.03\% &  0.06\% &   0.02\% &     \\
SQ       &   0.02\% &  0.01\% &      &     \\
NP-LOC   &   0.02\% &  0.00\% &   0.02\% &  0.04\% \\
NP-TTL   &   0.02\% &  0.01\% &   0.02\% &  0.01\% \\
NP-HLN   &   0.02\% &  0.02\% &   0.01\% &  0.00\% \\
SINV     &   0.01\% &  0.00\% &      &     \\
FRAG     &   0.01\% &  0.01\% &   0.00\% &     \\
WHADVP   &   0.01\% &  0.00\% &   0.01\% &  0.18\% \\
LST      &   0.01\% &     &      &     \\
WHADJP   &   0.00\% &  0.00\% &   0.00\% &     \\
X        &   0.00\% &  0.00\% &   0.00\% &  0.00\% \\
SBARQ    &   0.00\% &  0.00\% &      &     \\
NP-MNR   &   0.00\% &  0.00\% &   0.00\% &     \\
NP-CLR   &   0.00\% &     &   0.00\% &  0.02\% \\
INTJ     &   0.00\% &  0.00\% &   0.00\% &  0.01\% \\
NP-TPC   &   0.00\% &  0.00\% &   0.00\% &  0.00\% \\
NP-VOC   &   0.00\% &     &   0.00\% &  0.00\% \\
NP-DIR   &   0.00\% &     &   0.00\% &     \\
ADVP|PRT &      &     &      &  0.00\% \\
\bottomrule
\end{tabular}
    \caption{Label distribution for the detailed chunking tagset. Empty cells indicate that the combination of chunking label and phrase label is not present in the training data. Cells rounded to 0.00\% indicate labels that are exceptionally rare in the training data.\label{tab:chunk:labels:detailed}}
    
\end{subtable}

\begin{subtable}[b]{.5\textwidth}
    \centering\footnotesize
\begin{tabular}{lrr}
\toprule
{} &  original &  sampled \\
\midrule
S        &     39.20\% &    24.99\% \\
VP       &     25.37\% &    16.97\% \\
NP       &     22.66\% &    15.06\% \\
PP       &      5.35\% &     5.11\% \\
SBAR     &      2.65\% &     3.54\% \\
SINV     &      2.51\% &     3.50\% \\
ADJP     &      0.68\% &     2.59\% \\
ADVP     &      0.38\% &     2.30\% \\
QP       &      0.37\% &     2.24\% \\
FRAG     &      0.29\% &     2.26\% \\
UCP      &      0.13\% &     2.12\% \\
SQ       &      0.09\% &     2.11\% \\
WHNP     &      0.08\% &     2.06\% \\
NX       &      0.08\% &     2.08\% \\
SBARQ    &      0.05\% &     2.11\% \\
PRN      &      0.02\% &     2.05\% \\
WHADVP   &      0.02\% &     2.10\% \\
NAC      &      0.02\% &     2.04\% \\
CONJP    &      0.01\% &     1.44\% \\
WHPP     &      0.01\% &     0.92\% \\
X        &      0.01\% &     1.31\% \\
RRC      &      0.00\% &     0.57\% \\
INTJ     &      0.00\% &     0.22\% \\
LST      &      0.00\% &     0.05\% \\
ADVP|PRT &      0.00\% &     0.00\% \\
WHADJP   &      0.00\% &     0.25\% \\
PRT|ADVP &      0.00\% &     0.00\% \\
PRT      &      0.00\% &     0.01\% \\
\bottomrule
\end{tabular}
    \caption{Label distribution for LCA prediction. The left column indicates the original distribution of labels for token pairs for the whole training data. The right column indicates the relative frequency of labels for sampled token pairs}
    \label{tab:label-distribution-lca}
\end{subtable}

\begin{subtable}{.5\textwidth}
    \centering\footnotesize
    \begin{tabular}{c|c|c|c|c|c}
         & B & I & E & S & PCT  \\\hline
         & 43.3\% & 14.1\% & 18.1\% & 10.7\% & 13.8\%   
    \end{tabular}
    \caption{\texttt{B}: beginning, \texttt{I}: inside, \texttt{E}: end, \texttt{S}: Single, \texttt{PCT}: punctuation: punctuation is not considered for evaluation}
    \label{tab:chunk:labels:simple}
\end{subtable}

\caption{Label distributions for the different probing tasks}

\end{table}

\newpage

\subsection{Chunking results}\label{app:chunk:res}
\begin{table}[H]
    \centering
\begin{subtable}{\textwidth}
\centering
\begin{tabular}{cl|ll||ll||ll}
\toprule
    \multirow{2}{*}{task} & \multirow{2}{*}{train/test} & \multicolumn{2}{c}{DistilBERT} & \multicolumn{2}{c}{BERT} & \multicolumn{2}{c}{XLNet}\\
    & & task & sel. & task & sel. & task & sel.\\\hline
    \multirow[t]{5}{*}{chunking simple} & orig./orig. & 94.6 & 15.6 & 94.0 & 15.5 & 94.0 & 20.6 \\
         & $.33$/orig. &  93.9 &   17.5 & 92.6 & 16.9 & 92.9 & 20.8  \\
         & $.33$/$.33$ &  92.2 &   21.9 & 91.0 & 21.5 & 91.0 & 24.5  \\
         & $.67$/orig. &  92.9 &   20.5 & 92.1 & 20.2 & 91.7 & 23.5  \\
         & $.67$/$.67$ &  90.8 &   27.0 & 89.9 & 28.5 & 88.8 & 25.7  \\\hline
    \multirow[t]{5}{*}{chunking detailed} & orig./orig. &  89.8 & 11.9 & 89.3 & 13.9  & 89.1 & 19.4 \\
         & $.33$/orig.  &  88.4 &  12.7 & 86.6 & 13.6 & 87.5 & 19.0 \\
         & $.33$/$.33/$ &  85.2 &  19.4 & 84.5 & 21.5 & 84.8 & 22.2 \\
         & $.67$/orig.  &  85.0 &  18.9 & 83.1 & 19.5 & 83.9 & 20.1 \\
         & $.67$/$.67$  &  81.5 &  23.2 & 79.2 & 22.8 & 80.2 & 21.8 \\\hline
    \end{tabular}
\caption{Results for chunking experiments with DistilBERT, BERT and XLNet\label{tab:full:results:chunking}
}
\end{subtable}

\begin{subtable}{\textwidth}
    \centering\footnotesize
\begin{tabular}{lrrrr}
\toprule
{} &      B &     I &     E &     S \\
\midrule
B &  19999 &   228 &    12 &   189 \\
I &    310 &  6141 &   277 &    40 \\
E &     22 &   154 &  8652 &   105 \\
S &    184 &    41 &   122 &  4530 \\
\bottomrule
\end{tabular}
    \caption{Confusion matrix for chunking experiments with RoBERTa.    \label{fig:conf-matrix-chunk}
}
\end{subtable}

\begin{subtable}{\textwidth}
    \centering\footnotesize
    \includegraphics[width=.99\textwidth]{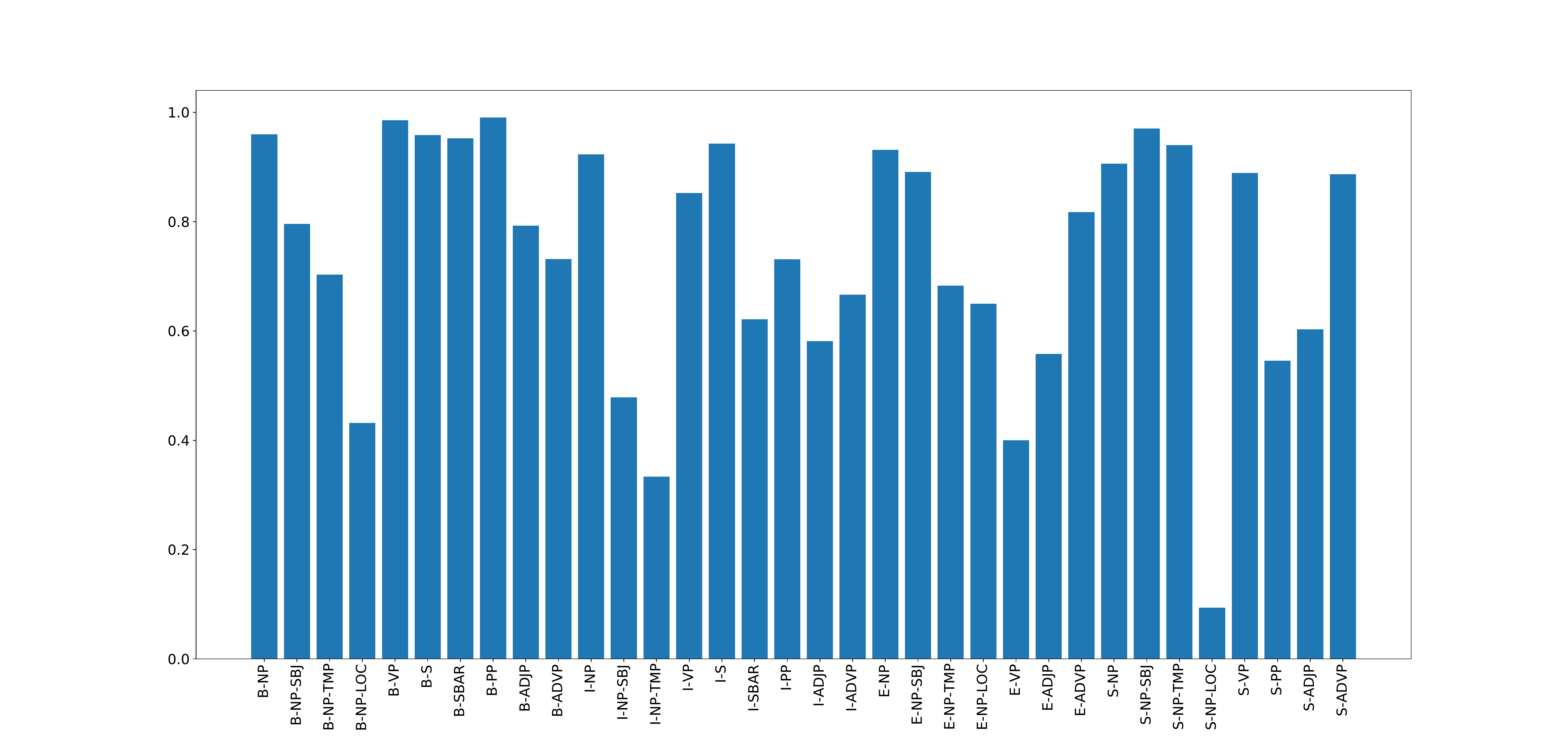}
    \caption{Accuracy per category for the most frequent labels in the detailed tagset. All beginnings of frequent constituents are recognized with high accuracy. The classifier is also able to distinguish between different kinds of NPs, such as \texttt{NP} without further specification, subject NPs (\texttt{NP-SBJ}) or temporal and local NPs (\texttt{NP-TMP},\texttt{NP-LOC})\label{tab:conf-matrix-chunk-detailed-b}}
\end{subtable}


    \caption{Results for chunking}
    \label{tab:chunk:full:results}
\end{table}

\newpage

\subsection{LCA prediction results}\label{app:lca:res}
\begin{figure}[H]
    \centering
\begin{subtable}{\textwidth}
\centering\small
\begin{tabular}{cl|ll||ll||ll||ll}
    \multirow{2}{*}{task} & \multirow{2}{*}{train/test} & \multicolumn{2}{c}{RoBERTa} & \multicolumn{2}{c}{DistilBERT} & \multicolumn{2}{c}{BERT} & \multicolumn{2}{c}{XLNet}\\
    & & task & sel. & task & sel. & task & sel. & task  & sel. \\\hline
    \multirow[t]{5}{*}{$concat$} & orig./orig. &  84.4 & 59.9 & 83.0 & 57.2 & 83.0 & 60.6 & 83.3 & 68.0 \\
         & $.33$/orig. & 81.0 & 62.1 & 83.5 & 57.0 & 81.1 & 61.3 & 80.9 & 67.7 \\
         & $.33$/$.33$ & 78.9 & 56.0 & 80.8 & 54.6 & 79.4 & 56.5 & 80.8 & 62.7 \\
         & $.67$/orig. & 78.6 & 60.7 & 77.8 & 57.4 & 77.3 & 63.0 & 76.3 & 56.3 \\
         & $.67$/$.67$ & 73.7 & 56.2 & 75.4 & 52.8 & 73.7 & 57.7 & 72.8 & 56.1 \\\hline
    \multirow[t]{5}{*}{\maxs} & orig./orig. & 69.3 & 42.7 & 68.6 & 40.1 & 63.4 & 39.7 & 70.5 & 54.5 \\
         & $.33$/orig. & 68.1 & 46.3 & 69.0 & 42.0 & 61.9 & 41.4 & 66.0 & 48.2 \\
         & $.33$/$.33$ & 66.5 & 45.2 & 66.4 & 36.2 & 60.3 & 39.6 & 60.1 & 43.2 \\
         & $.67$/orig. & 65.8 & 45.0 & 64.5 & 41.6 & 61.0 & 43.6 & 59.2 & 37.8 \\
         & $.67$/$.67$ & 62.3 & 43.9 & 63.3 & 38.0 & 59.0 & 39.4 & 55.1 & 32.0 \\\hline
    \multirow[t]{5}{*}{$avg$} & orig./orig. & 63.9 & 36.9 & 66.8 & 38.8 & 62.3 & 38.4 & 57.5 & 42.3 \\
         & $.33$/orig. & 64.3 & 37.2 & 66.1 & 38.7 & 63.9 & 39.3 & 58.9 & 45.0 \\
         & $.33$/$.33$ & 62.0 & 36.8 & 64.1 & 31.3 & 61.8 & 37.1 & 54.8 & 34.5 \\
         & $.67$/orig. & 63.4 & 37.2 & 63.4 & 36.5 & 62.9 & 41.7 & 57.3 & 38.4 \\
         & $.67$/$.67$ & 59.6 & 36.5 & 61.3 & 34.0 & 59.4 & 37.2 & 51.5 & 27.1\\\hline
    \end{tabular}
\caption{LCA prediction results. The performance gains of $concat$ wrt. $avg$ and \maxs are not matched by higher performance in the control task for $concat$. Thus $concat$ shows not only the best task performance but also the highest selectivity for LCA prediction.}
\label{tab:full:results:lca}
\end{subtable}

\vspace{10pt}

\begin{subtable}{\textwidth}
\centering					
\footnotesize
\begin{tabular}{lrrrrrrr}
\toprule
{} &    NP &    VP &     S &  SBAR &   PP &  ADJP &  ADVP \\
\midrule
NP   &  3269 &   167 &   289 &    18 &   83 &    13 &    15 \\
VP   &   307 &  3631 &   295 &    41 &   79 &     5 &    31 \\
S    &   258 &   162 &  5830 &    56 &   25 &     4 &    12 \\
SBAR &     8 &    11 &     7 &   218 &    8 &     0 &     0 \\
PP   &    61 &    31 &    13 &     7 &  827 &     8 &     4 \\
ADJP &    24 &    16 &     2 &     0 &    0 &    71 &     6 \\
ADVP &     2 &     4 &     2 &     1 &    0 &     0 &   120 \\
\bottomrule
\end{tabular}
    \caption{Confusion matrix for LCA prediction the most frequent constituents labels for RoBERTa when trained and evaluated on non-manipulated data. The columns represent predicted values, rows represent actual values. Some categories are better represented in the probing classifiers than others. For example, prepositional phrases are recognized quite reliably, but adjectival phrases are confused for \texttt{VPs} and \texttt{NP}s in a number of cases. \texttt{NPs} are frequently confused with all other categories. The reason might be that a variety of different phenomena are collected under \texttt{NP}, such as appositions and relative clauses.\label{tab:conf-matrix-lca}						    
}
\end{subtable}

\vspace{10pt}

\begin{subtable}[b]{.35\textwidth}							
\centering			
\small
    \begin{tabular}{l|cc}									
    \toprule									  
     & left & right  \\		
    \midrule
	RoBERTa & 66.5 & 40.8 \\
	DistilBERT & 68.0 & 43.6 \\
	BERT & 64.8 & 40.9 \\
	XLNet & 62.1 & 40.7\\
   \bottomrule
    \end{tabular}
    \caption{Results for single-token baseline on LCA prediction }
\label{tab:results:single-token-baseline}						    
\end{subtable}

\begin{subfigure}[b]{.6\linewidth}
    \centering
    \includegraphics[width=.9\linewidth]{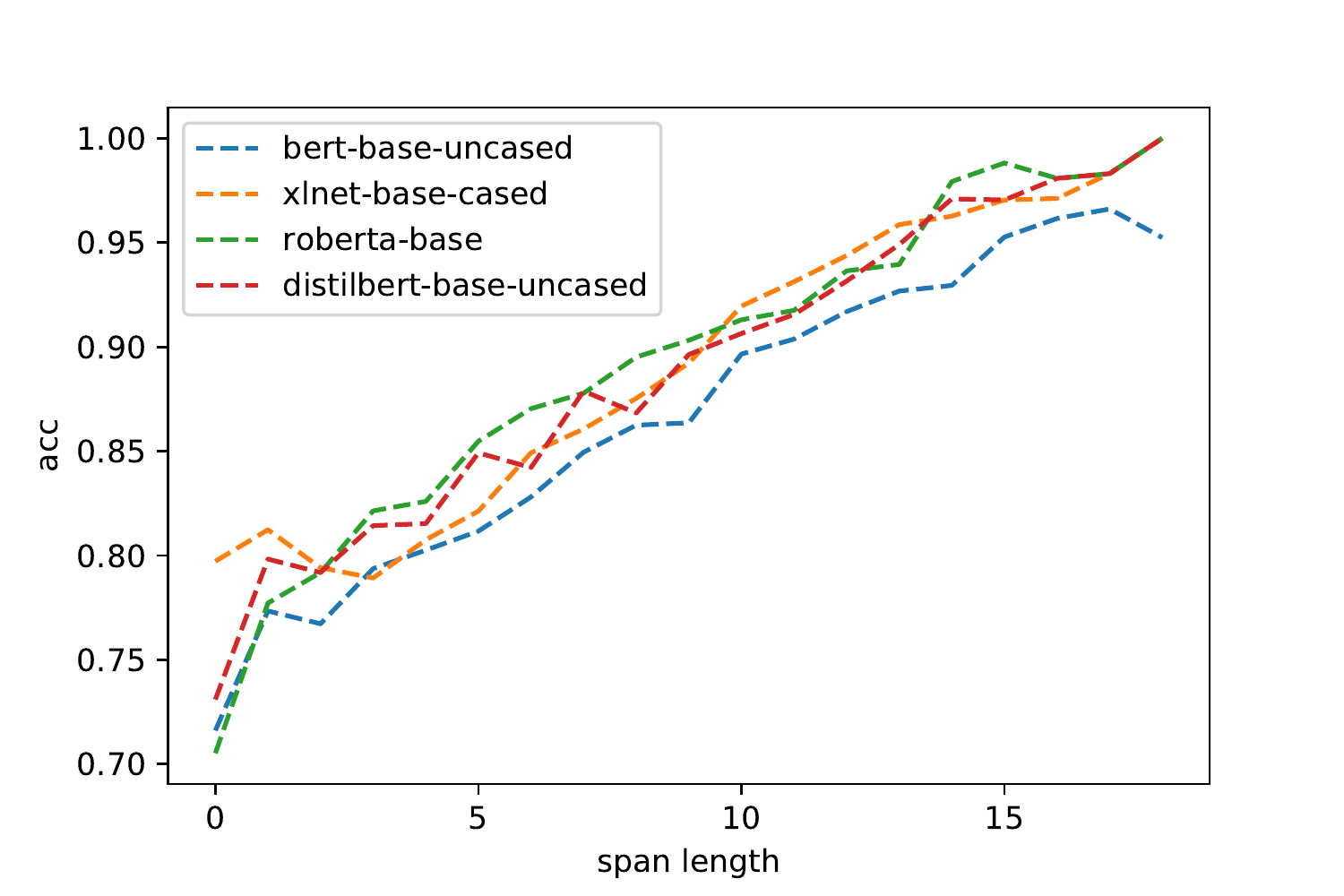}
    \caption{For LCA prediction, all models are more accurate when the distance between two tokens is higher}
    \label{fig:acc-per-len}
\end{subfigure}
    \caption{Results for LCA prediction}
    \label{tab:lca:full:results}
\end{figure}

\newpage

\subsection{Neuron-level results for LCA prediction}\label{app:neuron:lca}

\begin{figure}[h]
    \centering
    \includegraphics[width=.55\linewidth]{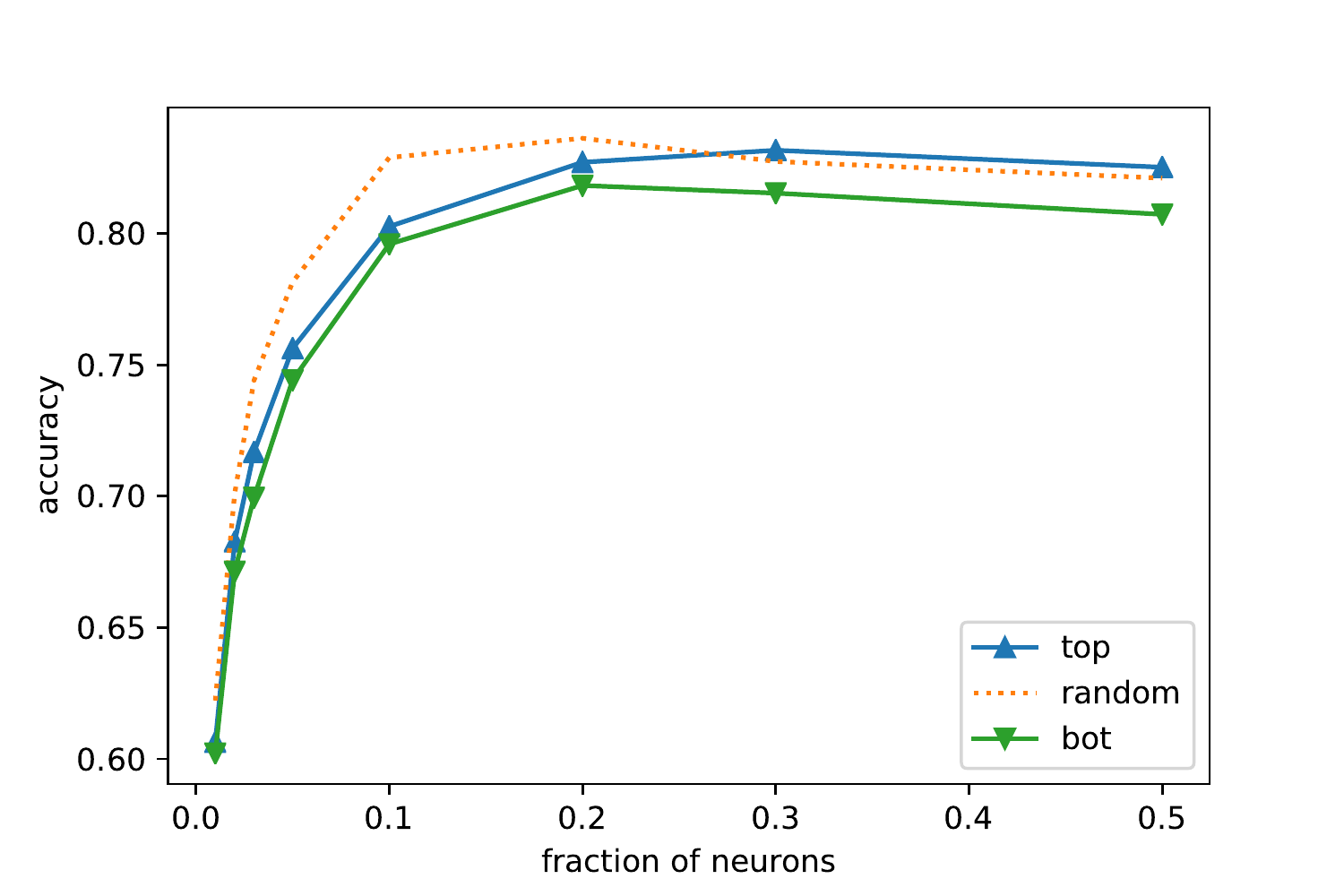}
    \caption{Accuracy for RoBERTa probes when trained on fractions of top, random or bottom neurons identified by linguistic correlation analysis \citep{durrani-individual:emnlp20}. The classifiers are trained on LCA prediction, using concatenated representations. Fractions on the x-axis refer to the dimensions of concatenated representations. Thus, the absolute number of considered neurons is twice as high as in the respective plots for chunking in Fig.~\ref{fig:acc-min-neurons}.}
    \label{fig:min:neurons:lca}
\end{figure}

\subsection{Neuron orderings for different datasets}\label{fig:overlap-datasets}

\begin{table}[h]
    \centering
    \small
\begin{tabular}{l|rrr}
\toprule
{\% neurons} &  orig. \& .33 &  orig. \& .67 &  .33 \& .67 \\
\midrule
1\% &           21.2\% &           16.2\% &         15.2\% \\
2\% &           18.1\% &           15.6\% &         23.1\% \\
3\% &           17.4\% &           12.7\% &         20.4\% \\
5\% &           19.8\% &           16.6\% &         21.2\% \\
10\% &           25.4\% &           20.7\% &         26.6\% \\
20\% &           35.7\% &           29.6\% &         34.4\% \\
30\% &           42.6\% &           38.1\% &         43.7\% \\
50\% &           59.1\% &           56.4\% &         59.4\% \\
\bottomrule
\end{tabular}
    \caption{Overlap of fractions of top neurons for chunking (simple tagset) when classifiers are trained on different datasets. For each dataset, a neuron ranking list is obtained. This table shows the size of the fraction of neurons that are ranked among the most salient $x\%$ of neurons for two different datasets. For example, $20.7\%$ of the $10\%$ most salient neurons for the original dataset are also among the $10\%$ most salient neurons for the .67 dataset.} 
    
\end{table}

\subsection{Results for parse tree reconstructions from all language models}\label{app:results:parsetrees}

\begin{table*}[h]
    \centering
    \small
\begin{tabular}{l|rrrrr}
\toprule
{} &  RoBERTa &  BERT &  DistilBERT &  XLNet &  Random BERT \\
\midrule
orig./orig.     & 82.58 &  80.42 &        79.88 &     80.52 &        51.36 \\
.33/orig. &         80.88 &              77.73 &                    77.65 &             78.99 &        49.60 \\
.33/.33  &         77.84 &              73.97 &                    74.09 &             75.23 &        46.13 \\
.67/orig. &         78.30 &              74.63 &                    74.81 &             75.37 &        47.95 \\
.67/.67  &         72.77 &              69.63 &                    69.72 &             69.91 &        44.71 \\
\bottomrule
\end{tabular}
    \caption{Labeled F1 scores for all datasets and models for reconstructing full parse trees}
\end{table*}

\subsection{Comparing parse trees reconstructed from different models}\label{app:comp:parsetrees}

\begin{table}[H]
    \centering
    \small
\begin{tabular}{l|p{1.8cm}p{1.8cm}p{1.8cm}p{1.8cm}p{2.2cm}p{1.8cm}}
\toprule
{} &  RoBERTa vs. XLNet &  RoBERTa vs. BERT &  BERT vs. XLNet &  DistilBERT vs. BERT &  DistilBERT vs. RoBERTa & DistilBERT vs. XLNet \\
\midrule
orig.     &                          81.83 &                           82.80 &                               80.39 &                                      83.17 &                                 82.34 &                                     80.19 \\
.33/orig. &                          79.71 &                           80.21 &                               77.60 &                                      81.60 &                                 80.53 &                                     77.88 \\
.33/.33  &                          76.02 &                           76.52 &                               73.99 &                                      78.72 &                                 76.78 &  74.30 \\
.67/orig. &                          76.54 &                           77.08 &                               74.07 &                                      78.68 &                                 77.78 &                                     74.40 \\
.67/.67  &                          71.75 &                           72.83 &                               69.56 &                                      75.01 &                                 73.45 &                                     69.89                                    \\ 
\bottomrule
\end{tabular}
    \caption{Labeled F scores for comparing constituent trees predicted by different models against each other. The comparison of trees predicted by DistilBERT and BERT yields the highest F scores, hence trees reconstructed from these models are most similar.}
    \label{tab:comparing-full-trees-results}
\end{table}

\begin{table}[H]
    \centering
    \small
\begin{tabular}{lrrrr}
\toprule
{} &  RoBERTa &  BERT &  DistilBERT \\
\midrule
BERT       &     0.59 &  &      \\
DistilBERT &     0.57 &  0.64 &         \\
XLNet      &     0.52 &  0.52 &        0.51 \\
\bottomrule
\end{tabular}
    \caption{Pearson correlation of sentence-level F scores for different LM0 on original data. The correlations between F scores are lowest when comparing XLNet to BERT, RoBERTa and DistilBERT. The correlations are highest between the latter three models.}
    \label{tab:my_label}
\end{table}

\newpage

\subsection{Examples for reconstructed parse trees}

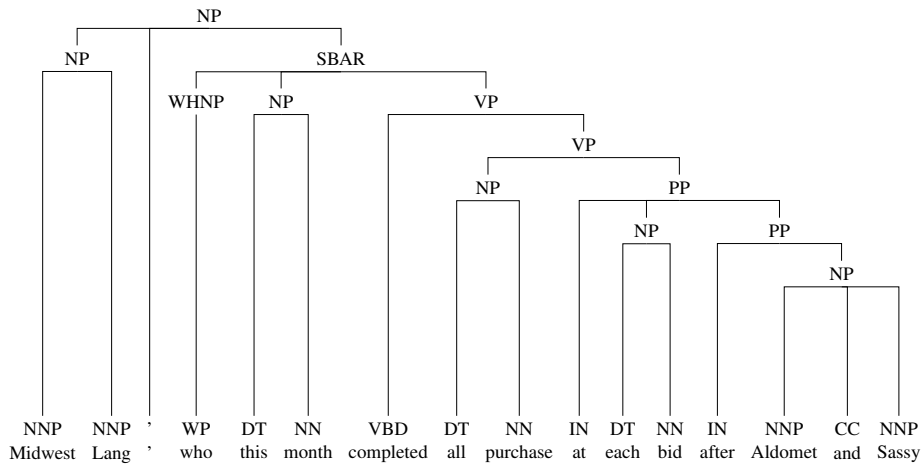
\begin{figure}[H]\centering
\begin{tikzpicture}
\tikzset{level distance=25pt}
\tikzset{every tree node/.style={align=center,anchor=north}}
\tikzset{frontier/.style={distance from root=240pt}}
\tikzset{edge from parent/.style= {draw,edge from parent path={(\tikzparentnode.south) -| (\tikzchildnode)}}}

\begin{scope}[scale=0.65, shift={(0pt,200pt)}]
\node at (-60pt,10pt) {Gold tree from PTB};
\Tree [.NP-SBJ
        [.NP NP\\Dale NNP\\Lang ]
        ,\\, 
        [.SBAR
          [.WHNP WP\\who ]
          [.S
            [.NP-TMP DT\\this NN\\week ]
            [.VP
              VBD\\completed
              [.NP
                [.NP DT\\the NN\\acquisition ]
                [.PP
                  IN\\of
                  [.NP
                    [.NP DT\\the NN\\publisher ]
                    [.PP
                      IN\\of
                      [.NP NNP\\Ms. CC\\and NNP\\Sassy ] ] ] ] ] ] ] ]
         ]
\end{scope}

\begin{scope}[scale=0.65, shift={(0pt,-100pt)}]
\node at (-60pt,10pt) {Roberta prediction};
\Tree [.NP
        [.NP NNP\\Dale NNP\\Lang ]
        ,\\,
        [.SBAR
          [.WHNP WP\\who ]
          [.NP DT\\this NN\\week ]
          [.VP
            VBD\\completed
            [.NP
              [.NP DT\\the NN\\acquisition ]
              [.PP
                IN\\of
                [.NP
                  [.NP DT\\the NN\\publisher ]
                  [.PP
                    IN\\of
                    [.NP NNP\\Ms. CC\\and NNP\\Sassy ] ] ] ] ]
           ] ] ]
\end{scope}

\begin{scope}[scale=0.65, shift={(0pt,-400pt)}]
\node at (-60pt,10pt) {RoBERTa .67:};
\Tree [.NP
        [.NP NNP\\Midwest NNP\\Lang ]
        ,\\,
        [.SBAR
          [.WHNP WP\\who ]
          [.NP DT\\this NN\\month ]
          [.VP
            VBD\\completed
            [.VP
              [.NP DT\\all NN\\purchase ]
              [.PP
                IN\\at
                [.NP DT\\each NN\\bid ]
                [.PP
                  IN\\after
                  [.NP NNP\\Aldomet CC\\and NNP\\Sassy ] ] ] ] ]
           ] ]
\end{scope}

\end{tikzpicture}
    \caption{Example for reconstructing parse trees: The subject phrase of sentence 23 from the development set. In the RoBERTa prediction, the SBAR and S nodes of the relative clause are conflated to an SBAR node. In the RoBERTa prediction with .67 of the tokens replaced, the short chunks are correctly recognized, but the VP of the relative clause is structured differently.}
    \label{fig:dev23:reconstructed}
\end{figure}

\begin{figure}\centering
\begin{tikzpicture}
\tikzset{level distance=25pt}
\tikzset{every tree node/.style={align=center,anchor=north}}
\tikzset{frontier/.style={distance from root=200pt}}
\tikzset{edge from parent/.style= {draw,edge from parent path={(\tikzparentnode.south) -| (\tikzchildnode)}}}

\begin{scope}[scale=0.65, shift={(0pt,200pt)}]
\node at (-60pt,10pt) {Gold tree from PTB};

\Tree [.S
        [.NP-SBJ DT\\The NN\\company ]
        [.VP
          VBD\\had
          [.VP
            VBN\\sought
            [.NP
              [.NP NNS\\increases ]
              [.VP
                VBG\\totaling
                [.NP
                  [.NP [.QP \$\\\$ CD\\80.3 CD\\million ] ]
                  ,\\,
                  CC\\or
                  [.NP CD\\22 NN\\\% ] ] ] ] ] ]
         ]
\end{scope}

\begin{scope}[scale=0.65, shift={(0pt,-120pt)}]
\node at (-60pt,20pt) {RoBERTa orig:};

\Tree [.S
        [.NP DT\\The NN\\company ]
        [.VP
          VBD\\had
          [.VP
            VBN\\sought
            [.NP
              [.NP NNS\\increases ]
              [.VP
                VBG\\totaling
                [.QP \$\\\$ CD\\80.3 CD\\million ]
                ,\\,
                CC\\or
                [.NP CD\\22 NN\\\% ] ] ] ] ]
         ]
\end{scope}

\begin{scope}[scale=0.65, shift={(0pt,-450pt)}]
\node at (-60pt,20pt) {RoBERTa .67: };
\Tree [.S
        [.NP DT\\Half NN\\company ]
        [.VP
          VBD\\did
          [.VP
            VBN\\rejected
            [.NP
              [.NP NNS\\efforts ]
              [.VP
                VBG\\surrounding
                [.NP
                  [.QP \$\\\$ CD\\656.5 CD\\million ]
                  ,\\,
                  CC\\\&
                  [.NP CD\\124.5 NN\\yen ] ] ] ] ] ]
        .\\. ]
\end{scope}

\end{tikzpicture}
    \caption{Example for reconstructing parse trees: Sentence 4 from the development set. For the RoBERTa prediction, the unary leaf chain NP $\to$ QP is not predicted because it cannot be predicted by our version of the sequence labeling algorithm. Apart from that, the \textit{or}-NP is not recognized as a single NP. In the RoBERTa prediction for the sentence with .67 of the tokens replaced, the VP is recognized correctly even though there is an agreement mismatch (\textit{did rejected}). Diverging from the prediction for original data, the \textit{or}-NP is reconstructed correctly}
    \label{fig:dev4:reconstructed}
\end{figure}
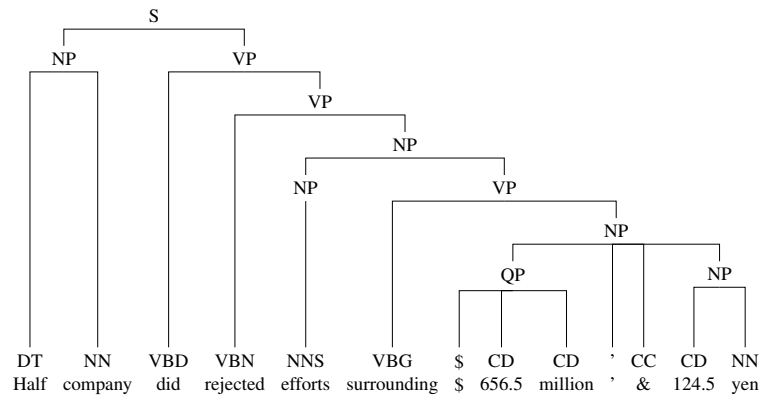

\newpage

\subsection{Computing infrastructure and experiment runtime}

All experiments where run on an Nvidia Titan XP GPU.
For a 12-layer model, the full set of chunking experiments takes 4 hours. This includes extracting the full neuron activation values for the training and evaluation data, all control task experiments and all experiments on the linguistic task.
The full set of LCA prediction experiments for a 12-layer model takes around 15 hours when using all ways of combining input representations ($concat$, $avg$, \maxs). 
Compared to the chunking experiments, an additional step is combining the activation values of token pairs. 
The full set of experiments for reconstructing parse trees takes around 3 hours when using a 12-layered language models and all datasets. 
Experiments with the 6-layer model DistilBERT take half the time.
The reported time includes time where the GPU itself is not active, e.g. times where representations are combined and written to harddrive. 
All experiments were powered with electricity from renewable energy sources.

\end{document}